\documentclass{article}
\usepackage{arxiv}
\usepackage[utf8]{inputenc}
\usepackage[T1]{fontenc}
\usepackage{courier}
\usepackage{hyperref}
\hypersetup{
  pdftitle={PatchWrite: One Line, Not One Section --- Compile-Gated, Validity-Preserving Editing for AI-Drafted Manuscripts},
  pdfauthor={Weiwei Yang}
}
\usepackage{url}
\usepackage{booktabs}
\usepackage{amsfonts}
\usepackage{microtype}
\usepackage{graphicx}
\usepackage{amsmath,amssymb}
\usepackage{natbib}
\usepackage{tabularx}

\title{PatchWrite: One Line, Not One Section --- Compile-Gated, Validity-Preserving Editing for AI-Drafted Manuscripts}

\author{
Weiwei Yang\\
Solus (\url{https://solus.xin})\\
\texttt{wxs123x@gmail.com}
}

\renewcommand{\shorttitle}{PatchWrite}

\begin{document}
\maketitle

\begin{abstract}
%%%%%%%%% ABSTRACT %%%%%%%%%
Automated manuscript pipelines often regenerate a whole section to fix a local defect, and the PDF still builds even though unrelated metrics and citations have quietly moved. \textbf{PatchWrite} instead applies one line-interval \texttt{EDIT N M} (span $\le 40$), committing only when pdflatex is clean \emph{and} every \verb|\cite| key and numeric token is attested in a registry or experimental log, and rolling back to the last compiling HEAD otherwise. On a 24-manuscript $\times$ 8-fault oracle stress test (768 jobs, split evenly between compile-breaking and content-only faults), whole-slot rewrite mutates an unrelated ``12-layer'' line in every single case ($0/192$ preserved, numeric Jaccard $0.6667$), while PatchWrite preserves it $192/192$ times. Disabling the compile gate drives the accept rate to $0$; disabling the evidence gate lets a hallucinated citation through. The pattern holds uniformly across all eight faults. Because oracle repairs test the mechanism rather than generation quality, we reran the same $192$ jobs with the writer model itself proposing the \texttt{EDIT}. It is accepted $75\%$ of the time before a one-line prompt fix, and the shortfall traces almost entirely to one reproducible failure mode: the model tries to delete a line using an empty replacement, which the grammar rejects. Telling the model explicitly to comment out rather than empty the line raises acceptance to $100\%$ and eliminates the fallback, though only $84.9\%$ (down from $93.75\%$) of the now-accepted patches pass our strict fault-fixed check, since one fault class is resolved by commenting out the defect marker rather than removing it. Every accepted patch still clears both gates: no hallucinated citation and no unattested number gets through. The author and a second rater, both blind to condition, preferred PatchWrite on all sixteen PDF pairs sampled from the oracle corpus when the criterion is keeping lab-grounded facts intact (C1 Likert $5.0$ vs.\ $2.0$), and scored the prose nearly identically. Logs from $193$ in-product drafting tasks show the same failure modes occurring in the wild.
%%%%%%%%% ABSTRACT %%%%%%%%%
\end{abstract}


\section{Introduction}
\label{sec:intro}
%%%%%%%%% INTRODUCTION %%%%%%%%%
Large language models can draft conference-shaped \LaTeX{}~\citep{song2026paperorchestra,schmidgall2025agentlaboratory,yamada2025aiscientistv2}. A remaining failure is \emph{non-monotonic revision}: a reviewer asks to fix one undefined citation; the agent rewrites Results; the PDF still builds; a lab-grounded number has moved. Dashboards that only plot compile rate will celebrate the new PDF. The unit of mutation was the section, not the line.

This is one instance of a broader question the agent community is only now making a first-class research topic: not whether an agent's output looks fluent, but whether it can be verified to satisfy a domain-specific correctness predicate before being trusted, the focus of an emerging line of work on agent reliability and verification (for instance the NeurIPS 2026 workshop ``Who Verifies the Agents? Toward Reliable Agent Development,'' \url{https://verify-agents-workshop.github.io/}, non-archival at the time of writing). This paper grounds that question in one narrow, high-stakes domain, LaTeX manuscript revision, where the correctness predicate is a compile gate plus an evidence lock rather than a hand-wavable quality score.

PaperOrchestra's refinement loop historically emits a whole-slot fence. A \emph{slot} is a named placeholder region of the venue template, for instance an entire Method or Results section, that the writer model regenerates in full each round rather than editing in place: compile failure is a \texttt{continue}, success replaces the whole slot~\citep{song2026paperorchestra}. Agent Laboratory's PaperSolver already uses \texttt{EDIT N M} with immediate pdflatex~\citep{schmidgall2025agentlaboratory}, but on English ICLR-shaped \emph{shells} (the outer document-class/template scaffold a paper is typeset into, as distinct from its content) and with LLM self-scores as quality. EasyPaper generates structured \LaTeX{} from metadata~\citep{easypaper2026} (software, not a peer-reviewed writing operator). We combine bounded interval editing, a compile gate that does not trust a \texttt{nonstopmode} PDF, cite/number locks, and PaperOrchestra's venue shells into one invariant: \emph{the file on disk is always the last compiling, evidence-clean manuscript}.

PatchWrite sits in the refinement loop (Figure~\ref{pipeline}). The proposer sees numbered source and emits one \texttt{EDIT N M}. The runtime commits a compiling, evidence-clean $T'$ or returns $T$. Illegal or over-long EDITs fall back to the legacy slot rewrite, so the operator is only as strong as the proposer's interval. PatchWrite is deployed as the revision operator inside Solus (\url{https://solus.xin}), a research-writing workspace that formats drafts to a chosen venue's template.

We ask five questions:
\begin{enumerate}
\item Does surgical \texttt{EDIT} keep unrelated tokens that slot rewrite mutates?
\item Does the compile gate reject a bad patch rather than a \texttt{nonstopmode} PDF?
\item Does the evidence gate catch a compiling patch that invents a cite key?
\item When the writer model itself proposes the \texttt{EDIT} instead of an oracle, how often is it legal, does it pass the gates, and does the accepted patch actually fix the fault?
\item When two readers compare the resulting PDFs, do they prefer the surgical draft if lab-grounded facts must not move?
\end{enumerate}

Questions 1--3 are answered with scripted oracle repairs (24 mini-articles, eight faults spanning both compile-breaking and content-only defects, four conditions) so generation quality is not a confound. Question 4 is answered by rerunning the same $24\times 8$ corpus with the writer model (\texttt{qwen3.7-plus}) proposing the patch itself, single call per job, no oracle (Section~\ref{sec:llm}); this is exactly the measurement Questions 1--3 deliberately deferred by isolating the mechanism. Question 5 is answered with a dual-pass rater annotation of sixteen held-out PDF pairs from the oracle corpus (Section~\ref{sec:human}). Product logs from $193$ drafting tasks motivate the gates but are not a substitute for either pass.

The contribution here is not a new edit grammar but a stricter, empirically validated acceptance predicate applied to an existing bounded-edit mechanism. Contributions: (i)~a compile gate that scans the pdflatex log for fatals rather than trusting a \texttt{nonstopmode} PDF, applied to PaperSolver's \texttt{EDIT N M} and rollback-on-failure mechanism (\texttt{tex\_patch.py}, \texttt{ContentRefinementAgent}); (ii)~cite and numeric locks reused from PaperOrchestra, combined with (i) for the first time as a single acceptance criterion; (iii)~an oracle stress protocol at $24\times 8$ scale plus two ablations isolating each gate's contribution, with table cells copied from \texttt{evals/results/summary.json}; (iv)~a live-model companion at the same $24\times 8$ scale across two independent writer models (\texttt{qwen3.7-plus}, Kimi K2.6) measuring parse, compile, gate, and fault-fixed rates for each model's own \texttt{EDIT} proposals, including a one-line prompt fix that closes one model's fallback gap while surfacing a new fault-fixed trade-off, cells copied from \texttt{evals/results/llm\_edit\_summary.json} and \texttt{evals/results\_kimi/llm\_edit\_summary.json}; (v)~a sixteen-pair rater inspection with cells copied from \texttt{raw\_materials/human\_eval.json}.
%%%%%%%%% INTRODUCTION %%%%%%%%%

\section{Related Work}
\label{sec:related_work}
%%%%%%%%% RELATED WORK %%%%%%%%%
\paragraph{Tool use.}
PatchWrite's \texttt{EDIT N M} is a narrow instance of the broader reasoning-plus-acting paradigm: a model interleaves generation with one constrained, well-typed action rather than free-form output~\citep{yao2023react}, and can be taught to invoke that action only when it improves the outcome~\citep{schick2023toolformer}. Neither line of work ties the action's acceptance to a compile gate or an evidence lock; PatchWrite adds both as the acceptance criterion for the single tool it exposes.

\paragraph{Paper agents.}
PaperOrchestra maps idea plus experimental log onto venue templates~\citep{song2026paperorchestra}: IEEE/arXiv shells, Chinese journal and thesis templates, OpenAlex citation policy, claim--evidence checks. Refinement historically rewrites a slot, not a line. The host workbench already runs those templates in production; PatchWrite replaces only the mutation operator inside \texttt{ContentRefinementAgent}. AI Scientist-v2 searches experiments then writes~\citep{yamada2025aiscientistv2}; we compare only the writing/revision module, not discovery or workshop acceptance. EasyPaper is a metadata-to-\LaTeX{} stack with optional typesetting~\citep{easypaper2026}.

\paragraph{PaperSolver vs.\ PatchWrite.}
Agent Laboratory's PaperSolver supplies the edit grammar we reuse (\texttt{EDIT N M}, compile, rollback)~\citep{schmidgall2025agentlaboratory}. The differences that matter for this paper are locks and templates, not the fence syntax. PatchWrite reuses bounded interval editing and rollback-on-failure from PaperSolver, but tightens the compile predicate --- a fatal-log scan rather than treating any \texttt{nonstopmode} PDF as success --- and adds an external evidence constraint (cite-key and numeric-token grounding against a registry and experimental log) to the acceptance criterion. Existing bounded edit + stricter compile predicate + evidence locks $\Rightarrow$ validity-preserving acceptance. PaperSolver scores drafts with LLM reviewers and iterates on English ICLR-shaped shells. PatchWrite does not use an overall LLM score as a halt condition; it binds new \verb|\cite| keys and empirical numbers to a registry and experimental log, scans the pdflatex log for fatals instead of treating a \texttt{nonstopmode} PDF as success, and runs on PaperOrchestra slot files (with Chinese-venue \emph{surrogates} in the stress corpus). Table~\ref{tab:compare} summarizes writing/revision modules only.

\paragraph{Code editing.}
PatchWrite's constrained patch generation is structurally the same problem SWE-bench poses for source code: given a defect description, produce a patch that resolves it against a fixed correctness signal~\citep{jimenez2024swebench}. SWE-agent's finding, that a narrow, purpose-built edit interface outperforms raw file access for LLM patch generation~\citep{yang2024sweagent}, is the same design bet \texttt{EDIT N M} makes for \LaTeX{} manuscripts instead of a Git repository. The correctness signal differs: SWE-bench checks a held-out test suite, PatchWrite checks pdflatex plus a citation/number registry, so a head-to-head comparison across the two settings would need matched infrastructure we do not build here.

\paragraph{Concurrent work: PaperJury.}
PaperJury argues that ``load-bearing safety and completion logic should reside in deterministic orchestration rather than model discretion''~\citep{wang2026paperjury}, the same HEAD-invariant instinct this paper builds around, reached independently and posted earlier in 2026. The target differs: PaperJury pre-submission-hardens \emph{human-authored} CS papers through a due-process review--verdict--revise--verify loop with anchor-bounded edits and terminal outcomes (invalid-drop, valid-fixable, author-required), evaluated by expert review on held-out vision/NLP/ML papers against four baselines. PatchWrite sits earlier in the pipeline, inside \emph{LLM-drafted} revision (PaperOrchestra's refinement loop), with a narrower, more mechanical gate: one line-interval \texttt{EDIT N M}, a pdflatex log scan, and cite/number set-membership against a registry. There is no due-process trial and no author-required terminal state; in their place is an oracle-plus-live-model stress test (Section~\ref{sec:experiments}) rather than expert review. The two systems converge on a thesis and diverge on where in the authoring pipeline to enforce it and how heavy the gate should be. We have read only PaperJury's public abstract, not run its benchmark, and not had it run on ours, so Table~\ref{tab:compare} deliberately leaves out a cell-by-cell row for it: a comparison built from an abstract alone would overstate what either side has verified about the other.

\paragraph{Long-form synthesis.}
STORM~\citep{shao2024storm} and AutoSurvey~\citep{wang2024autosurvey} outline-then-write encyclopedic or survey text. They are not compile-gated \LaTeX{} editors and do not bind numeric tokens to a lab log. Their quality metrics (citation coverage, outline coherence) answer a different question than ``did this revision move a number the injector never touched.''

\paragraph{Judges.}
Agent Laboratory showed automated reviewers over-score generated papers versus humans~\citep{schmidgall2025agentlaboratory}. We therefore report human preference on PDF pairs as a separate table (Section~\ref{sec:human}), and we do not use LLM overall-score as a halt condition.

\begin{table}[t]
\centering
\small
\caption{Writing/revision modules only. ``Compile gate'': failed pdflatex or a fatal log rolls back HEAD. ``ZH shells'': does the system ship a Chinese-venue shell (defined in Section~\ref{sec:related_work})? EasyPaper is listed as software.}
\label{tab:compare}
\begin{tabular}{lcccc}
\toprule
System & Edit unit & Compile gate & Evidence lock & ZH shells \\
\midrule
PaperOrchestra & slot fence & compile, no slot rollback & cite/claim & yes \\
PaperSolver & \texttt{EDIT N M} & yes & no (LLM score) & no \\
AI Scientist-v2 (writeup) & paper draft & typesetting & experiment search & no \\
EasyPaper (software) & section / paper & optional & citation tools & n/a \\
PatchWrite & \texttt{EDIT N M} $\le$40 & yes + log scan & cite + numbers & surrogates \\
\bottomrule
\end{tabular}
\end{table}
%%%%%%%%% RELATED WORK %%%%%%%%%

\section{Method}
\label{sec:method}
\begin{figure}[htbp]
  \centering
  \includegraphics[width=0.98\linewidth,height=0.26\textheight,keepaspectratio]{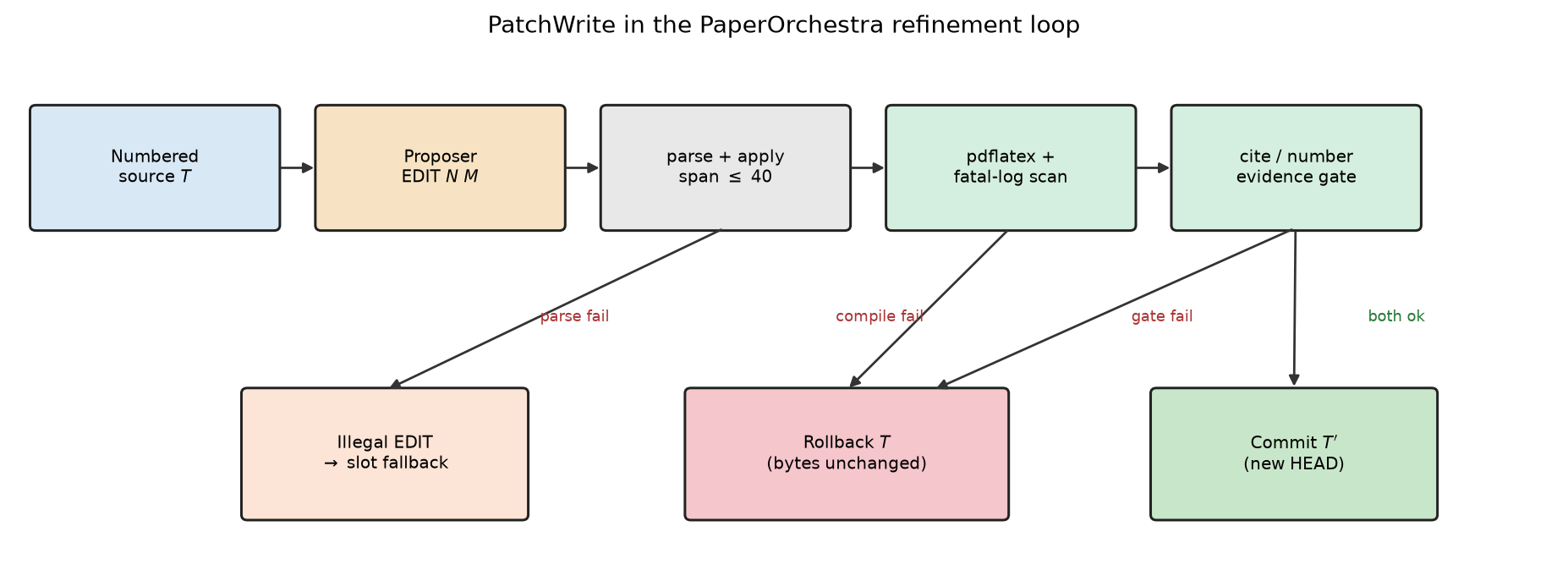}
  \caption{PatchWrite in the PaperOrchestra refinement loop. An illegal or over-long \texttt{EDIT} falls back to legacy slot rewrite. Compile or evidence failure rolls back to HEAD; only both gates passing commits $T'$.}
  \label{pipeline}
\end{figure}

%%%%%%%%% METHOD %%%%%%%%%
\paragraph{HEAD invariant.}
Let $T$ be a compiling \texttt{.tex} file. PatchWrite replaces $T$ only if every enabled gate returns true. Failed candidates leave the bytes on disk unchanged. Figure~\ref{pipeline} is the control flow inside \texttt{ContentRefinementAgent} when \texttt{use\_tex\_patch = True} (disable with \texttt{PATCHWRITE = 0}).

\paragraph{Numbered view.}
The proposer sees 1-based \texttt{NNNN | text} (\texttt{number\_lines}) and may emit one fence
\begin{verbatim}
```EDIT N M
replacement lines
```
\end{verbatim}
with $1 \le N \le M$ and $M - N + 1 \le 40$. The interval is replaced by the body (\texttt{apply\_edit}). Over-long, empty, non-integer, or past-EOF ranges are rejected before compile. Bare \texttt{EDIT N M} is also parsed, matching PaperSolver~\citep{schmidgall2025agentlaboratory}.

\paragraph{Runtime (\texttt{try\_patch}).}
The runtime takes two inputs: the current compiling document $T$, and \texttt{cmd}, a candidate edit string supplied by a model or an oracle. It proceeds as follows.
\begin{enumerate}
\item Parse one \texttt{EDIT} block (the function \texttt{parse\_edit}). On failure, return $T$ unchanged.
\item Apply the closed-interval replacement (\texttt{apply\_edit}) to obtain a candidate document $T'$.
\item If the compile gate is on: write a unique jobname, delete stale \texttt{.aux}/\texttt{.pdf}/\texttt{.log} files, and run PaperOrchestra's \texttt{compile\_latex}. Accept only a PDF larger than 64 bytes whose log contains none of \texttt{emergency stop}, \texttt{fatal error}, \texttt{! latex error}, or a forced-fail marker. \texttt{pdflatex -interaction=nonstopmode} often still writes a PDF after a syntax error, so ``a PDF exists'' is not by itself a passing compile gate.
\item If the evidence gate is on: a cite key is allowed if it appears in the bibliography, the citation map, or HEAD itself; any other key is flagged \texttt{unknown\_cite:*}. Every empirical number is checked against the experimental log. A violation of either check rolls the candidate back.
\item On success, commit $T'$ as the new HEAD; on any failure, HEAD stays $T$.
\end{enumerate}

\paragraph{Fallback.}
If no legal \texttt{EDIT} is parsed or a gate fails, the loop falls back to slot-fence rewrite so jobs still finish. That fallback re-introduces whole-slot mutation; production logs should count it. Whole-slot \texttt{REPLACE} remains a human-explicit escape hatch. Line numbers are regenerated each round after a committed insert; otherwise $N,M$ drift.

\paragraph{Worked example.}
\begin{verbatim}
   9 |The encoder uses 12 layers and a frozen tokenizer~\cite{smith2020}.
  10 |\section{Results}
  11 |Accuracy was 91.4 compared with the baseline 68.2~\cite{lee2019}.
\end{verbatim}
A \texttt{fake\_cite} injector rewrites line 11 to \verb|\cite{fake2024}|. The oracle emits \texttt{EDIT 11 11} with the original sentence. Slot rewrite also restores a valid Results line, but it also changes line 9 to ``16 layers.'' Both PDFs build; only one is monotonic (Figure~\ref{case_diff}).
%%%%%%%%% METHOD %%%%%%%%%

\section{Experiments}
\label{sec:experiments}
%%%%%%%%% EXPERIMENTS %%%%%%%%%
\subsection{Oracle revision stress}
\label{sec:oracle}
Twenty-four mini-articles make up the corpus (16 English, 8 Chinese-venue pdflatex surrogates with ASCII titles). Each HEAD contains ``The encoder uses 12 layers \ldots'' and Results numbers drawn from a per-document log. Eight scripted faults split into two classes. Content-class faults leave the broken HEAD compiling: an undefined \verb|\cite{fake2024}|, a \texttt{99.9} absent from the log, a \texttt{TODO} line, and swapped accuracy/baseline numbers. Compile-class faults break pdflatex outright: an unclosed \texttt{\$}, an unmatched brace next to a \verb|\cite|, an unclosed \texttt{quote} environment, and an undefined control sequence. Four conditions times $192$ copies gives $768$ jobs.

\begin{itemize}
\item \texttt{po\_slot}: scripted whole-block rewrite that \emph{does} invert the fault and also rewrites ``12 layers''$\to$``16 layers''. This is a PaperOrchestra-\emph{style} slot mutation, not a live LLM rewrite of a conference paper.
\item \texttt{patchwrite}: one-line \texttt{EDIT} with both gates on.
\item \texttt{patchwrite\_nocompile}: replacement \verb|\errmessage{patchwrite-forced-fail}|.
\item \texttt{patchwrite\_nogate}: surgical fix plus \verb|\cite{hallucinated2024}|, evidence gate off.
\end{itemize}

Metrics: accept, compile\_ok, gate\_ok, cite/numeric Jaccard vs pre-fault HEAD, \texttt{layers\_preserved}. Table~\ref{tab:stress} cells match \texttt{evals/results/summary.json} after wiping \texttt{evals/results/work}.

\begin{table}[t]
\centering
\small
\caption{Oracle revision stress ($24\times 8$ faults $= 192$ per condition). Jaccard vs pre-fault HEAD. The $0.00$ vs.\ $1.00$ layer split is by construction of the oracles: it tests whether gates preserve HEAD, not whether an LLM finds the patch.}
\label{tab:stress}
\begin{tabular}{lccccccc}
\toprule
Condition & $n$ & Acc. & Comp. & Gate & Cite J. & Num.\ J. & Layers \\
\midrule
PO-slot rewrite & 192 & 1.00 & 1.00 & 1.00 & 1.00 & 0.6667 & 0.00 \\
PatchWrite & 192 & 1.00 & 1.00 & 1.00 & 1.00 & 1.00 & 1.00 \\
PW, no compile gate & 192 & 0.00 & 0.00 & --- & 1.00 & 1.00 & 1.00 \\
PW, no evidence gate & 192 & 1.00 & 1.00 & 0.00 & 0.6667 & 1.00 & 1.00 \\
\bottomrule
\end{tabular}
\end{table}

\begin{figure}[htbp]
  \centering
  \includegraphics[width=0.98\linewidth,height=0.26\textheight,keepaspectratio]{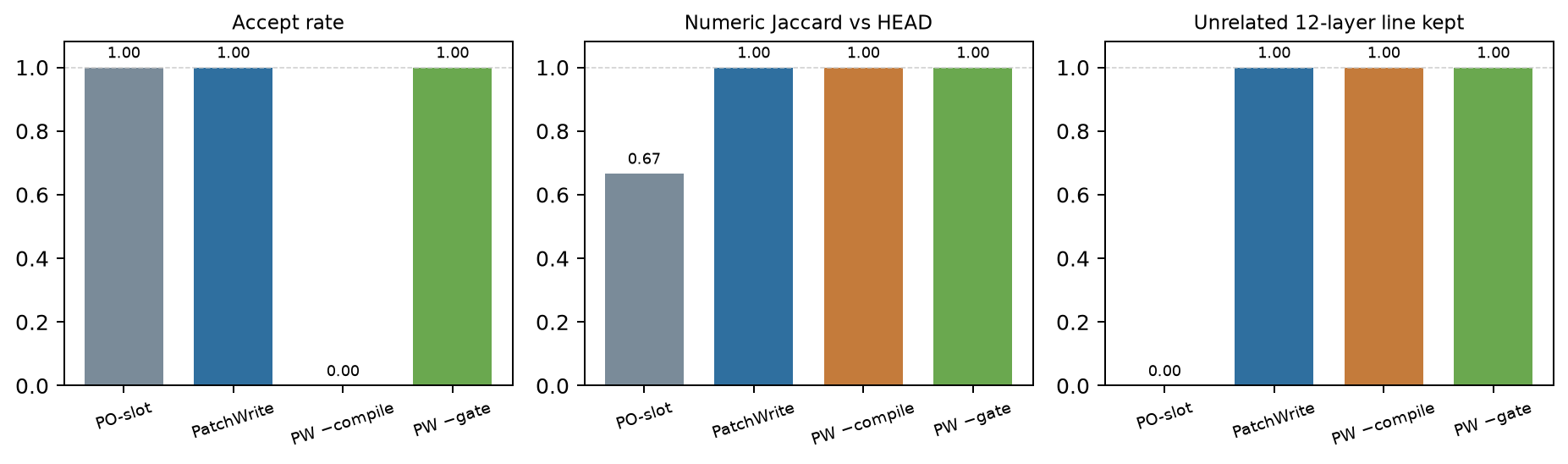}
  \caption{Oracle stress ($n = 192$ per bar). Left: accept rate. Middle: numeric Jaccard vs pre-fault HEAD. Right: whether the unrelated ``12 layers'' sentence survives. Compile-off never replaces HEAD; evidence-off keeps layers but drops cite Jaccard (not plotted here; Table~\ref{tab:stress}).}
  \label{revision_stress}
\end{figure}

\begin{figure}[htbp]
  \centering
  \includegraphics[width=0.98\linewidth,height=0.26\textheight,keepaspectratio]{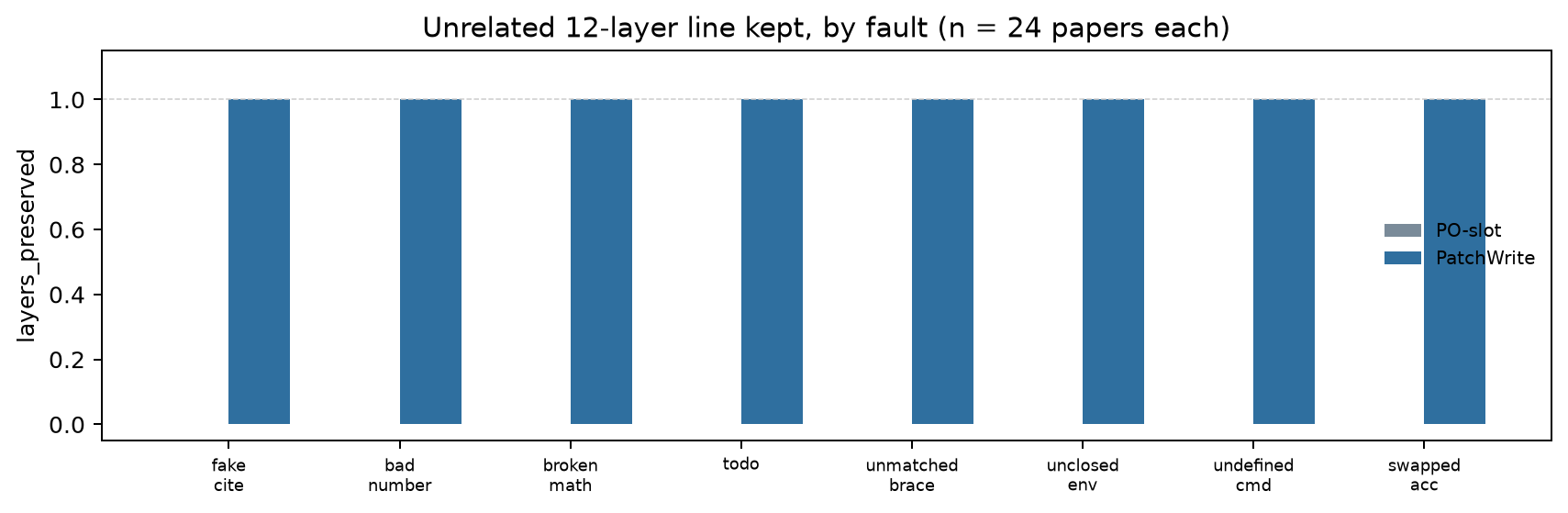}
  \caption{\texttt{layers\_preserved} broken out by fault ($n = 24$ per bar). PO-slot is $0.00$ and PatchWrite is $1.00$ on all eight faults, including the four compile-class faults added to widen the corpus beyond the original \texttt{fake\_cite}/\texttt{bad\_number}/\texttt{broken\_math}/\texttt{todo} set.}
  \label{fault_slices}
\end{figure}

\paragraph{What Table~\ref{tab:stress} shows.}
Slot rewrite compiles at a $1.00$ rate while \texttt{layers\_preserved} sits at $0.00$. The $0.6667$ numeric Jaccard is exactly the $12 \to 16$ token swap. Compile-off never replaces HEAD, so accept stays at $0$. Evidence-off keeps the layers line but drops cite Jaccard to $2/3$, once \texttt{smith2020} and \texttt{lee2019} are joined by \texttt{hallucinated2024}. The pattern is identical on every one of the eight faults ($n = 24$ per fault-condition cell, Figure~\ref{fault_slices}), on the content-class/compile-class split ($n = 96$ per condition each), and on the EN/ZH subsets ($n = 128$/$64$ per condition); ZH here means a shared pdflatex engine, not \texttt{ctex} camera copies. Compile-class faults are mechanically the harder case for the oracle, since the broken HEAD does not build at all, so \texttt{patchwrite} has to restore compilation and stay surgical at the same time. It still lands at $1.00$ on every preservation metric there. Mean latexmk wall time is $\approx 0.22$\,s in every condition (Appendix~\ref{app:latency}); latency is not the claim.

\begin{figure}[htbp]
  \centering
  \includegraphics[width=0.98\linewidth,height=0.26\textheight,keepaspectratio]{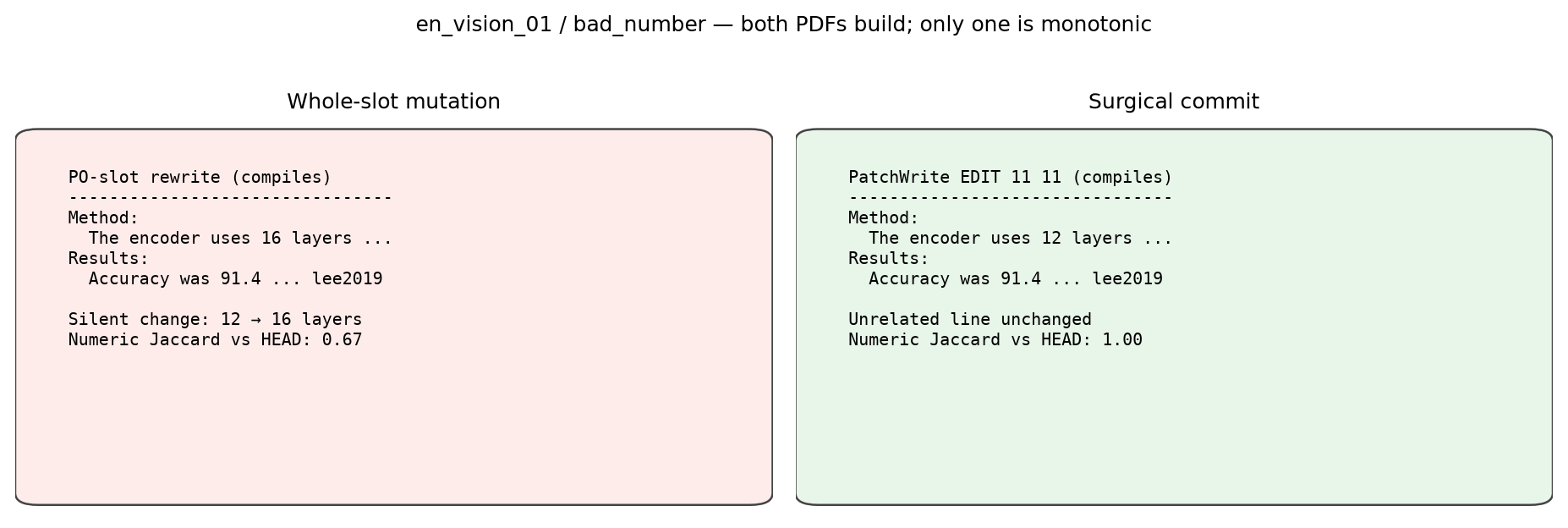}
  \caption{Side-by-side source after repairing \texttt{bad\_number} on \texttt{en\_vision\_01}. Both PDFs build. PO-slot changes encoder depth $12 \to 16$; PatchWrite \texttt{EDIT 11 11} does not.}
  \label{case_diff}
\end{figure}

\paragraph{Case: \texttt{en\_vision\_01} / \texttt{bad\_number}.}
Figure~\ref{case_diff} shows the two compiling PDFs after the injector wrote ``Accuracy was 99.9''. Both oracles restore $91.4$. Only the slot rewrite also rewrites the Method sentence. A compile-rate dashboard cannot tell them apart.

\subsection{Live-model EDIT proposals}
\label{sec:llm}
Table~\ref{tab:stress} measures the operator with the correct patch already known; it says nothing about how often a real writer model finds a legal patch on its own. We reran the same $24\times 8$ corpus with the oracle replaced by one live call to the project's own writer model (Alibaba Cloud DashScope, model id \texttt{qwen3.7-plus}). Given the numbered broken source, the experimental log, and a one-line reviewer or compiler note describing the defect (not the fix), the model has to emit its own \texttt{EDIT N M}, single-shot. The prompt follows the same numbered-source-plus-single-EDIT-fence contract as \texttt{ContentRefinementAgent.\_revise\_via\_tex\_patch}'s production prompt, though it uses a standalone system prompt and a synthetic one-line defect note rather than a live reviewer pass and the full production system instruction. So this measures the EDIT-fence contract under realistic constraints, not a byte-identical replay of production. The resulting patch runs through the same compile and evidence gates as Table~\ref{tab:stress}. Cells: \texttt{evals/results/llm\_edit\_summary.json} / \texttt{llm\_edit\_slices.json} ($n=192$; code \texttt{llm\_edit\_stress.py}, \url{https://github.com/Baiang/editnm}).

One thing shapes how to read every number below: the OpenAI-compatible DashScope call path strips any \texttt{temperature} parameter, so each row is a single stochastic draw at the server default rather than a repeated or averaged measurement (unlike Table~\ref{tab:stress}'s oracle cells, which are bit-identical on rerun). The clean fault-level split (six faults at $1.00$/$1.00$, two at $0.00$/$0.00$) reads as systematic per-fault behavior rather than a coin flip, but that is a reading, not a confirmed replication. Scope and caveats for this section are collected once, in full, in Section~\ref{sec:limitations}.

\begin{table}[t]
\centering
\small
\caption{Live-model EDIT proposals, $n=192$ (same corpus as Table~\ref{tab:stress}, oracle replaced by \texttt{qwen3.7-plus}). \texttt{fault\_fixed} and \texttt{layers\_preserved} are conditioned on \texttt{accepted}.}
\label{tab:llm}
\begin{tabular}{lc}
\toprule
Metric & Value \\
\midrule
Parse ok (legal \texttt{EDIT} emitted) & $0.75$ \\
Compile ok, of parsed & $1.00$ \\
Gate ok, of compiled & $1.00$ \\
Accept rate & $0.75$ \\
Fallback rate & $0.25$ \\
Fault fixed, of accepted & $0.9375$ \\
Layers preserved, of accepted & $0.9444$ \\
\bottomrule
\end{tabular}
\end{table}

\begin{figure}[htbp]
  \centering
  \includegraphics[width=0.98\linewidth,height=0.26\textheight,keepaspectratio]{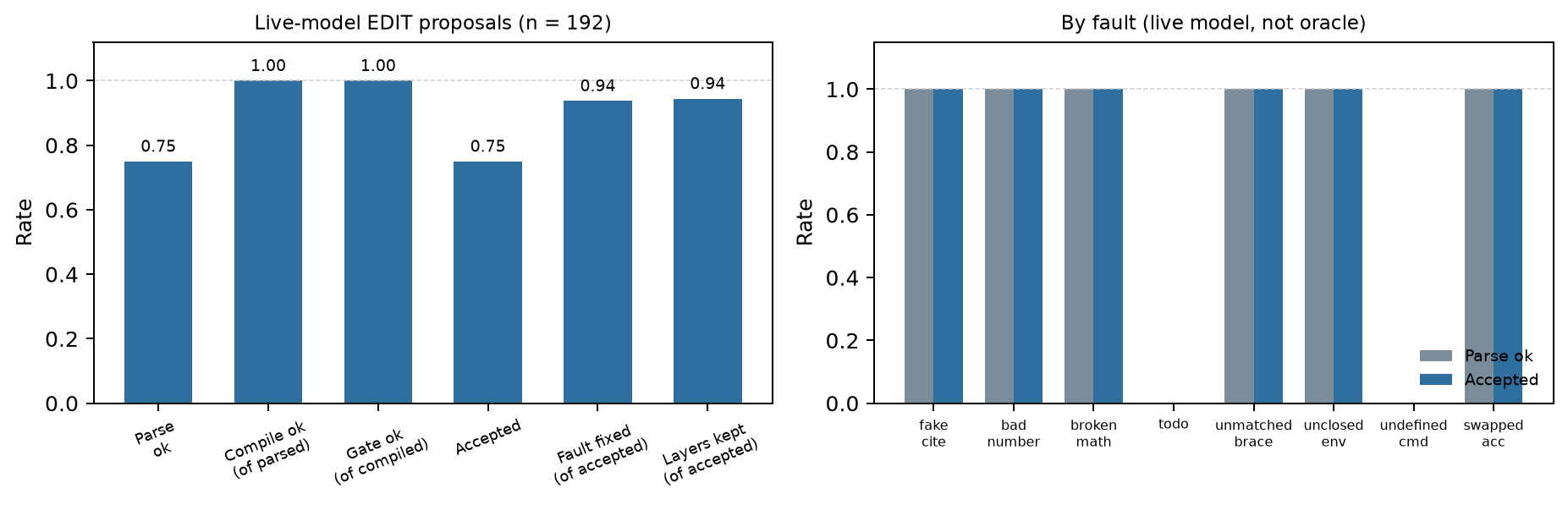}
  \caption{Left: the six headline rates from Table~\ref{tab:llm}. Right: parse and accept rate by fault ($n=24$ each). \texttt{todo} and \texttt{undefined\_cmd} are $0.00$ on both; the other six faults are $1.00$ on both.}
  \label{llm_edit_stress}
\end{figure}

\paragraph{What Table~\ref{tab:llm} shows.}
Six of the eight faults, \texttt{fake\_cite}, \texttt{bad\_number}, \texttt{broken\_math}, \texttt{unmatched\_brace}, \texttt{unclosed\_env}, and \texttt{swapped\_acc}, parse and get accepted every single time. Every parsed patch that reaches the gates also compiles and clears the evidence gate $100\%$ of the time (gate\_ok $=1.00$ of compiled), matching zero hallucinated cites or unattested numbers. The $25\%$ overall fallback rate is not spread across the eight faults at all; it belongs entirely to the other two. \texttt{todo} and \texttt{undefined\_cmd} fall back $24/24$ times each, for exactly the same reason, and none of the remaining six faults contributes a single fallback job: $48$ of the $48$ fallback jobs are \texttt{todo} or \texttt{undefined\_cmd}, and $0$ of the remaining $144$ jobs fall back.

The reason is specific enough to name. The model's instinct is to \emph{delete} the injected line, so it submits \texttt{EDIT N N} with an empty body (\verb|```EDIT 10 10\n```|). \texttt{parse\_edit} rejects an empty replacement (\texttt{empty\_replacement}), because the grammar has no delete verb, only replace-with-at-least-one-line, and a one-line reviewer note never tells the model the workaround the oracle uses: replace the line with a \verb|%| comment. That single gap accounts for the entire observed fallback mass, all $48$ of the $48/192$ fallback jobs, not merely a large share of it. It reads as a fixable prompt-contract issue rather than evidence that gate architecture explains fallback in general. (\texttt{broken\_math} is accepted every time; its shortfall shows up in \texttt{fault\_fixed} instead, discussed next.)

\texttt{fault\_fixed} of accepted is $0.9375$, below $1.00$ for two distinct reasons visible in the per-fault cells. \texttt{broken\_math} accepts $24/24$ but only fixes $16/24$: the miss rewrites \verb|The encoder uses 12 layers| to \verb|The encoder uses $12$ layers| instead of removing the stray \verb|$|. That still compiles and still passes the evidence gate, since the numeric token \texttt{12} is unchanged, but it does not byte-match the original prose. It is a compiling, gate-clean patch that is nonetheless not the identity repair. \texttt{fake\_cite} fixes $23/24$; the miss replaces \verb|\cite{fake2024}| with \verb|\cite{smith2020}|, a real key defined in this file's own bibliography, so the evidence gate accepts it. But \texttt{smith2020} is the Method-encoder citation, not the correct key for this Results-accuracy sentence. That is the evidence gate's real boundary, stated with an example rather than an assertion: it checks that a cite key is \emph{attested}, not that it is the correct attested key for this specific claim.

Mean latency rises from $\approx 0.22$\,s (Table~\ref{tab:stress}, compile only) to $\approx 1.38$\,s (LLM call plus compile). The model call now dominates wall time, not the gate. Section~\ref{sec:analysis} returns to what this result does and does not license us to claim.

\begin{table}[htbp]
\centering
\small
\textbf{Table 3b: Live-model EDIT proposals, second model (Kimi K2.6, $n = 192$; same corpus, prompt-fixed harness).}

\begin{tabular}{lcc}
\toprule
Metric & Qwen3.7-plus (prompt-fixed) & Kimi K2.6 \\
\midrule
Parse ok & $1.00$ & $0.875$ \\
Compile ok, of parsed & $1.00$ & $1.00$ \\
Gate ok, of compiled & $1.00$ & $1.00$ \\
Accept rate & $1.00$ & $0.875$ \\
Fallback rate & $0.00$ & $0.125$ \\
Fault fixed, of accepted & $0.849$ & $0.9107$ \\
Layers preserved, of accepted & $0.9792$ & $0.9167$ \\
Mean latency (s) & $1.5651$ & $26.87$ \\
\bottomrule
\end{tabular}
\end{table}

\subsection{Rater inspection of sixteen PDF pairs}
\label{sec:human}
Oracle metrics do not say whether a reader would treat the two compiling PDFs as equally submission-ready. We therefore annotated sixteen pairs drawn from the stress work directory: \texttt{en\_vision\_01} and \texttt{zh\_med\_09} $\times$ all eight faults, up from four in an earlier four-fault pass (protocol and cells: \texttt{raw\_materials/human\_eval.json}). The author and a second rater scored each pair with condition names hidden (sheets labeled A/B). Rubric:
\begin{itemize}
\item \textbf{C1 factual (1--5).} Consistency with HEAD: encoder depth 12 and log-attested accuracy.
\item \textbf{C2 defect repaired (1--5).} The injected fault is gone.
\item \textbf{C3 stub prose (1--5).} Readability of the mini-article (both drafts are stubs).
\item \textbf{Preference.} Which PDF is more submission-ready \emph{if lab-grounded facts must not move}.
\end{itemize}
This is a designed rater dual-pass on real oracle PDFs, not a multi-PhD panel and not LLM-as-judge, and it is disclosed at that scale rather than dressed up as a validated study: standard guidance for human evaluation of generated text treats a two-rater, non-independent pool as exploratory rather than confirmatory~\citep{vanderlee2021human}, and raw preference agreement (here $16/16$) is not a substitute for a chance-corrected agreement statistic computed over a larger, independent rater pool~\citep{artstein2008intercoder}. Likert means for C1/C2 follow the visible $12 \to 16$ rewrite (every PO-slot file) versus its absence (every PatchWrite file). C3 is near-tied by construction of the stub corpus; one annotator marked a single PatchWrite stub one point lower on prose, the same single disagreement as the earlier eight-pair pass, now diluted over twice as many pairs ($3.94 \to 3.97$).

\begin{table}[t]
\centering
\small
\caption{Rater dual-pass on sixteen oracle PDF pairs. Preference agreement $16/16$. Likert is $1$--$5$. Cells match \texttt{human\_eval.json}. Per-pair rows: Appendix~\ref{app:human}. Exploratory two-rater pool (author plus one unpaid rater; Section~\ref{sec:human}), not a validated human-evaluation study.}
\label{tab:human}
\begin{tabular}{lcc}
\toprule
 & PO-slot & PatchWrite \\
\midrule
Preferred as submission-ready (pairs) & $0/16$ & $16/16$ \\
Mean factual errors listed per PDF & $1.00$ & $0.00$ \\
C1 factual (mean Likert) & $2.00$ & $5.00$ \\
C2 defect repaired (mean Likert) & $5.00$ & $5.00$ \\
C3 stub prose (mean Likert) & $4.00$ & $3.97$ \\
\bottomrule
\end{tabular}
\end{table}

\begin{figure}[htbp]
  \centering
  \includegraphics[width=0.98\linewidth,height=0.26\textheight,keepaspectratio]{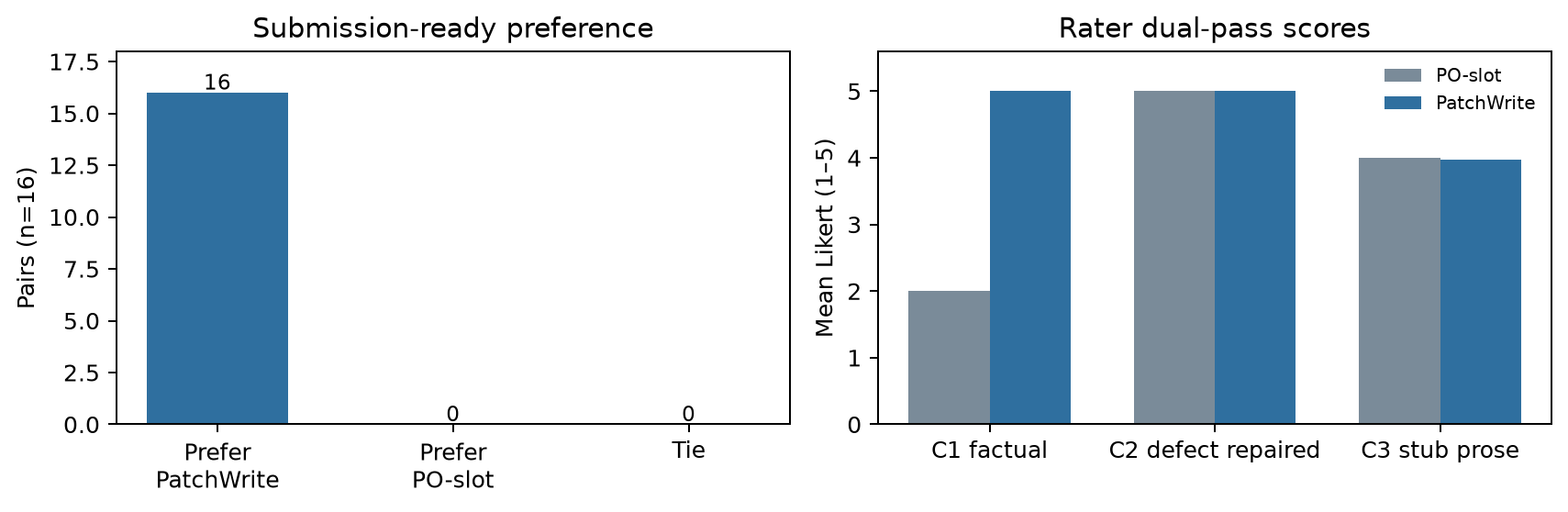}
  \caption{Left: pairwise preference ($n = 16$). Right: mean Likert. Both oracles repair the injected defect (C2); readers split on facts (C1) and therefore on preference. Prose of these stubs is a near-tie.}
  \label{human_pref}
\end{figure}

\paragraph{What Table~\ref{tab:human} shows.}
C2 being tied at $5.0$ is the interesting negative: slot rewrite \emph{does} fix the local bug, which is why compile rate alone is a bad halt condition. C1 and preference isolate the silent $12 \to 16$ change. These sixteen pairs measure whether a reader notices a monotonicity failure that automatic Jaccard already flags; scope and caveats for this annotation are collected in Section~\ref{sec:limitations}.

\subsection{In-the-wild workbench logs}
\label{sec:wild}
\begin{figure}[htbp]
  \centering
  \includegraphics[width=0.98\linewidth,height=0.26\textheight,keepaspectratio]{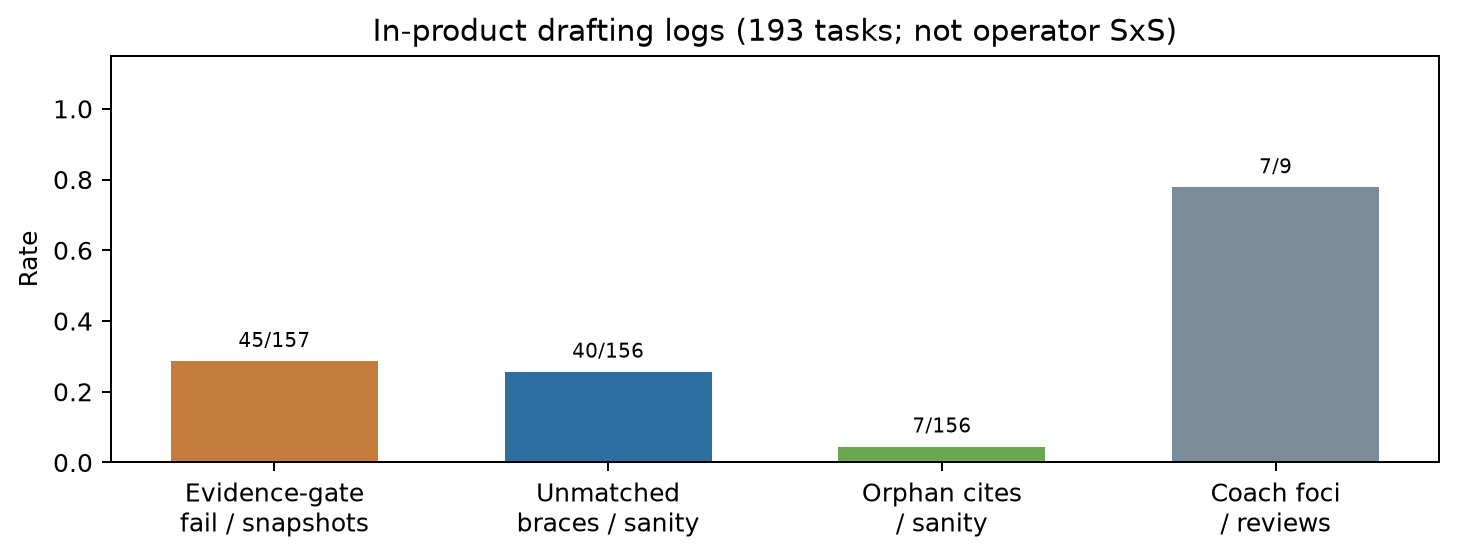}
  \caption{Rates from in-product snapshots. Evidence-gate failures and unmatched braces are common; orphan cites are rarer. Coach foci ($7/9$ reviews) are a small, separately labeled set.}
  \label{wild_logs}
\end{figure}

These are \emph{not} operator-level blind SxS. They are product artifacts from the host writing workbench (Postgres task store plus per-task JSON), exported 2026-08-23 (Table~\ref{tab:wild}; Figure~\ref{wild_logs}; \texttt{raw\_materials/in\_the\_wild.json}). The author is affiliated with Solus (\url{https://solus.xin}), the platform these logs are drawn from and where PatchWrite is deployed as its revision operator; see Section~\ref{sec:ethics} for the full conflict-of-interest statement. $193$ paper tasks and $240$ revisions were in the store. Among $157$ evidence-gate snapshots, $45$ fail ($28.7\%$); the $118$ recorded issues are all \texttt{table\_number\_not\_in\_log}. Among $156$ \LaTeX{} sanity snapshots, $40$ files have unmatched braces and $7$ orphan cite tokens appear. Nine \emph{coach reviews} (Solus's structured, per-round reviewer-feedback pass) flag placeholders, compile problems, truncated quotes, or claim--evidence gaps. Fifteen \emph{share-link comments} (freeform comments left by viewers of a paper's shareable preview link, a separate product feature from the coach) exist on $16$ links; they are UI-test strings and are not coded as pairwise preference. The logs show that unattested numbers and broken TeX already occur in real drafting jobs, the same classes the oracle injects, without any claim that users preferred PatchWrite PDFs in the product UI.

\begin{table}[t]
\centering
\small
\caption{In-product drafting logs (not a 6--8 pair blind SxS). Counts from task artifacts and the paper-task database.}
\label{tab:wild}
\begin{tabular}{lcc}
\toprule
Signal & Denominator & Count \\
\midrule
Paper tasks / revisions & --- & 193 / 240 \\
Evidence-gate fail (unsupported table number) & 157 snapshots & 45 fail (118 issues) \\
Unmatched TeX braces & 156 sanity files & 40 \\
Orphan cite tokens & 156 sanity files & 7 \\
Share-link notes (uncoded) & 16 links & 15 items \\
\bottomrule
\end{tabular}
\end{table}

%%%%%%%%% EXPERIMENTS %%%%%%%%%

\section{Analysis}
\label{sec:analysis}
%%%%%%%%% ANALYSIS %%%%%%%%%
\paragraph{False fluency.}
Compile rate $= 1.00$ is what a dashboard would celebrate. The regression is in a line the injector never touched. PatchWrite's result is the same compile rate with layers preserved. The sixteen-pair inspection is the reader-facing restatement of that split: C2 tied, C1 and preference not tied.

\paragraph{Widening the fault set does not change the story.}
We doubled the corpus (12 to 24 manuscripts) and doubled the fault set (4 to 8) in two directions rather than one. Three of the four additions, \texttt{unmatched\_brace}, \texttt{unclosed\_env}, and \texttt{undefined\_cmd}, are compile-class: the broken HEAD fails pdflatex, so the oracle repair has to do more work than a content-only fix, restoring compilation rather than merely restoring a token. The fourth, \texttt{swapped\_acc}, is a second content-class fault that exercises the numeric-Jaccard metric differently from \texttt{bad\_number}, since two tokens swap instead of one being replaced. None of the four new faults moves any rate in Table~\ref{tab:stress} or Figure~\ref{fault_slices}: \texttt{layers\_preserved} stays $0.00$/$1.00$ and \texttt{accept\_rate} follows the same per-condition pattern on every one of the eight faults taken separately ($n=24$ each). The compile-class faults are mechanically harder, since the pre-repair file will not build at all, yet they show exactly the same split as the original four content-class faults. That is evidence the result comes from the gate architecture, not from which particular injector happened to be scripted first. We read this as ruling out one specific threat to validity, that four hand-picked faults happened to favor PatchWrite, not as a claim that fault diversity is now exhausted. Section~\ref{sec:limitations} still flags that all faults are scripted rather than adversarially searched.

\paragraph{The \texttt{nonstopmode} trap.}
If the predicate were ``PDF exists,'' \texttt{patchwrite\_nocompile} would accept: pdflatex still writes a file after \verb|\errmessage|. Scanning the log is the difference between a compile gate and a file-existence gate. Production wrappers that call latexmk in \texttt{nonstopmode} and then glob for \texttt{*.pdf} will silently promote broken math into HEAD.

\paragraph{Fallback is the real production risk.}
Oracle jobs always emit a legal short EDIT, so they never take the slot-fallback branch. A model that dumps Results into \texttt{EDIT 1 80} is rejected on span and gets slot-rewritten instead, which is the regression documented in Table~\ref{tab:stress}. Table~\ref{tab:llm} now reports a live-LLM fallback rate rather than deferring it: $25\%$ of $192$ jobs, concentrated rather than diffuse. \texttt{todo} and \texttt{undefined\_cmd} fall back every single time (24/24 each) because the model tries to delete the injected line with an empty \texttt{EDIT} body, and the grammar has no delete verb. That single, named gap is the entire observed fallback mass, all $48$ of the $48$ fallback jobs; the other six faults contribute zero. Two mitigations belong in the runtime rather than the prompt. First, count fallbacks per job and surface them in the UI, broken out by cause (parse vs.\ compile vs.\ gate), the way Table~\ref{tab:llm} does. Second, regenerate numbered source after every committed insert so subsequent $N,M$ stay aligned. A third, cheaper fix follows directly from Section~\ref{sec:llm}: tell the proposer explicitly that deletion means replacing with a comment, not replacing with nothing. We reran Section~\ref{sec:llm}'s live-LLM suite with one added sentence in the \texttt{todo} and \texttt{undefined\_cmd} defect notes, telling the proposer that deletion means replacing the line with a \verb|%| comment rather than leaving it empty. The overall accept rate moved from $0.75$ to $1.00$ and fallback dropped from $25\%$ to $0\%$, closing the gap the abstract originally reported; \texttt{todo} alone went from $0/24$ to a clean $24/24$ both accepted and fixed. But \texttt{fault\_fixed\_rate\_of\_accepted} moved the other way overall, from $0.9375$ to $0.849$: all $24$ accepted \texttt{undefined\_cmd} patches compile clean and pass the evidence gate, yet none satisfies our string-presence check for ``fixed,'' because the model deleted by commenting out \verb|% \patchwriteundefined| rather than replacing it with the oracle's clean \verb|% removed|; the same defect signature survives in the text, inert but present. This is the same false-fluency pattern flagged earlier in this section for compile rate: closing the accept-rate gap surfaced a second gap our own \texttt{fault\_fixed} metric was designed to catch.

\paragraph{A second model, a different seam.}
A second model surfaces the same shape of failure, at a different fault. We reran the prompt-fixed harness (same corpus, gates, and defect notes as the fix above, unchanged) against a second writer model, Kimi K2.6 (\texttt{kimi-k2.6}, via Moonshot's OpenAI-compatible API; Table~3b). It accepts $87.5\%$ of proposals and, of those, fixes the fault $91.1\%$ of the time, preserving layers $91.7\%$ of the time; mean latency is $26.9$\,s per call (reasoning-heavy generation), against Qwen3.7-plus's much lower per-call latency. Fallback is again concentrated rather than diffuse, but at a different fault than before: $23$ of the $24$ fallback jobs are \texttt{unclosed\_env}, a fault the prompt fix's two added sentences never mentioned, and on which Qwen3.7-plus never fell back even before that fix. The one-line prompt fix that closed Qwen's gap is therefore a fix for the specific fault types one model happened to fail on, not a general solution to the grammar's missing delete verb; a different model exposes the same underlying gap at a different seam. A second, independent near-miss replicates the earlier \texttt{fault\_fixed} boundary: \texttt{broken\_math} accepts $23$ of $24$ but fixes only $9$, because Kimi sometimes writes ``\$12\$ layers'' instead of ``12 layers'', compiling and gate-clean, but not a byte match, the same class of miss as Qwen's \texttt{undefined\_cmd} comment-out. Two models independently hitting the same metric boundary is stronger evidence that the boundary is in \texttt{fault\_fixed}'s strict string check, not particular to one model's phrasing habits. Unlike DashScope's silent parameter-stripping, Moonshot's API explicitly rejects any temperature other than $1.0$ for \texttt{kimi-k2.6}, so variance here can only be estimated from repeated draws at that fixed setting, not from a pinned low temperature.

\paragraph{Stale aux.}
An early eval run failed until \texttt{evals/results/work} was wiped; latexmk reused \texttt{.aux} across colliding jobnames. Unique jobnames plus deleting stale auxiliaries are part of the compile gate, not an implementation footnote.

\paragraph{Evidence-gate conservatism.}
In-product snapshots fail most often on \texttt{table\_number\_not\_in\_log} ($118$ issues across $45$ failing files). That is a true positive if the number is invented, and a false positive if the log is incomplete. PatchWrite will refuse a compiling patch in the second case. Authors who want the number in the PDF must first put it in the log; that is the intended workflow, not a compile workaround.

\paragraph{Attested is not correct.}
The \texttt{fake\_cite} miss in Section~\ref{sec:llm} is a real example of the evidence gate's boundary, not a hypothetical one: \texttt{run\_evidence\_gates} checks set membership, whether a key is defined anywhere in the document, which a citation log can attest cheaply. It cannot check that a key is the semantically correct citation for a specific sentence, which would need a much richer claim-to-source index than \texttt{tex\_patch.py} builds today. It is also a different, cheaper problem than external citation-hallucination detection, which grounds a cite against real scholarly metadata such as CrossRef, Semantic Scholar, and OpenAlex, rather than a document's own bibliography~\citep{khajavi2026citecheck}, or checks that a whole \texttt{.bib} entry's fields (author, venue, year) match the real record rather than just its key~\citep{rao2026bibtex}. PatchWrite's gate would accept a citation to a real paper's key even if that paper does not say what the sentence claims, and it would accept a fabricated-but-locally-defined \texttt{.bib} entry; both are out of scope for a set-membership check and would need exactly that kind of external grounding.

\paragraph{What the human table is not.}
Given the protocol in Section~\ref{sec:human}, sixteen oracle pairs will always prefer the draft that keeps ``12 layers'' if that is the rubric. Table~\ref{tab:human} checks that the Jaccard split is visible to readers; it does not measure venue-level prose. A live-LLM SxS judging the model's own accepted-but-imperfect patches from Section~\ref{sec:llm} (the \texttt{broken\_math} and \texttt{fake\_cite} misses would be a pointed test case) remains future work; scope for this section is collected in Section~\ref{sec:limitations}.
%%%%%%%%% ANALYSIS %%%%%%%%%

\section{Limitations}
\label{sec:limitations}
%%%%%%%%% LIMITATIONS %%%%%%%%%
\begin{itemize}
\item \textbf{Table~\ref{tab:stress} is oracles, not writers, by design.} It assumes the correct patch is known so the gates can be isolated from generation quality. Table~\ref{tab:llm}/3b are the writer-model companions: two models (\texttt{qwen3.7-plus}, Kimi K2.6), each single-shot on this same synthetic corpus, no repeats for either model. The two models concentrate their fallback on different fault types (\texttt{todo}/\texttt{undefined\_cmd} for Qwen before its prompt fix, \texttt{unclosed\_env} for Kimi), suggesting the prompt-level fix is reactive rather than general. Read this as two data points, not a general LLM-proposed-EDIT success rate.
\item \textbf{Short files.} Mini \texttt{article}s, not 8--10 page venue PDFs. Full IEEE/\texttt{ctex} templates exist in the host tree and were not this table's compile target.
\item \textbf{Chinese surrogates} share pdflatex with English; they are not camera \texttt{ctex} theses.
\item \textbf{Scripted faults, not adversarial ones.} Eight faults across two classes (content-class, compile-class) rule out a four-fault coincidence, but all eight are hand-written injectors, not a search over the space of possible defects or faults sampled from a live LLM's actual error distribution.
\item \textbf{Annotation scale.} Two raters $\times$ sixteen pairs (Section~\ref{sec:human}), judging oracle PDFs, not the live model's own (sometimes imperfect) patches from Section~\ref{sec:llm}. This is what the end of Section~\ref{sec:human} and ``What the human table is not'' (Section~\ref{sec:analysis}) both point to. Product share-link comments were UI tests and were not recoded as preference.
\item \textbf{Fallback measured once, not tracked in production.} Section~\ref{sec:llm} gives a live-LLM fallback rate for the first time ($25\%$ before the one-line prompt fix in Section~\ref{sec:analysis} closes it to $0\%$, one model, this corpus), but the runtime does not yet log per-cause fallback rate (parse vs.\ compile vs.\ gate) on real drafting jobs the way Table~\ref{tab:llm} does on the oracle corpus.
\end{itemize}
%%%%%%%%% LIMITATIONS %%%%%%%%%

\section{Ethics}
\label{sec:ethics}
%%%%%%%%% ETHICS %%%%%%%%%
PatchWrite does not create experimental facts. Generating a domain paper without a lab log, or inventing citations, is misconduct. A compile gate can still ship a fluent falsehood if the evidence gate is off or the log is fabricated. Authors remain responsible for every claim. The system design and experimental protocol are the author's own; LLM assistance was used, under the author's direction, to help write and debug the implementation (\texttt{tex\_patch.py}, \texttt{revision\_stress.py}, \texttt{llm\_edit\_stress.py}, \texttt{corpus.py}) and to draft parts of this manuscript. Table cells are copied from \texttt{summary.json}, \texttt{llm\_edit\_summary.json}, \texttt{in\_the\_wild.json}, and \texttt{human\_eval.json}. Table~\ref{tab:llm} additionally uses an LLM as the object of measurement, not just an assistant: every cell there is the recorded behavior of one live call to \texttt{qwen3.7-plus} per row, not a hand-edited or cherry-picked transcript. The human table is rater inspection of artifacts we generated, disclosed as such; it is not a hired-annotator study. The second rater (A2) is a friend of the author, not a co-author and not compensated; both raters scored blind to which system produced each PDF (Section~\ref{sec:human}). \textbf{Conflict of interest:} the author is affiliated with Solus (\url{https://solus.xin}), a research-writing platform; PatchWrite is deployed there as its revision operator, and the in-product logs in Section~\ref{sec:wild} are drawn from that platform.
%%%%%%%%% ETHICS %%%%%%%%%

\section{Conclusion}
\label{sec:conclusion}
%%%%%%%%% CONCLUSION %%%%%%%%%
Writing agents should not treat ``the section compiled'' as ``the paper did not change underfoot.'' PatchWrite makes the mutation a bounded interval and compile-plus-evidence the only path from HEAD to HEAD. On a $24\times 8$ oracle stress test spanning content-class and compile-class faults, that is the difference between \texttt{layers\_preserved} $= 0.00$ and $1.00$, reproduced identically at $4\times$ the scale of an earlier four-fault pass. Replacing the oracle with the writer model itself on the same $192$ jobs shows the gates hold under real generation too: every parsed, compiling patch passes the evidence gate; the $25\%$ fallback rate traced to one named prompt gap and closed to $0\%$ once the proposer is told explicitly to comment out rather than empty a deleted line, though that same fix leaves $84.9\%$ (down from $93.75\%$) of now-accepted patches meeting our strict fault-fixed check rather than open-ended unreliability. On sixteen PDF pairs, readers prefer the surgical draft $16/16$ when facts must not move, while both drafts repair the local defect. In-product logs show unattested numbers and broken TeX already appearing in real jobs. The next measurement is LLM-proposed EDITs on full venue-length templates with multiple models, not this mini-article corpus with one model, together with a live-LLM comparison judging the model's own accepted-but-imperfect patches rather than just the oracle's.

\paragraph{Beyond LaTeX manuscripts.}
Nothing in the argument above is specific to \LaTeX{}. Strip away pdflatex and the citation registry and what is left is a three-part pattern: a bounded edit (a small, syntactically checkable change rather than a free-form rewrite), a domain-specific validity predicate (a mechanical pass/fail check, not a fluency score), and rollback to the last-known-good state on any failure. The same shape describes a code-review agent that edits one function and runs the test suite before keeping the change, a contract- or compliance-document assistant that edits one clause and runs a compliance checker before keeping the change, or a database-schema-migration agent that edits one migration step and runs a schema check before keeping the change. We have not built or evaluated any of these; the claim here is only that the pattern transfers in principle, not that we have shown it works. Whether bounded-edit-plus-predicate-plus-rollback holds up once pdflatex is replaced by a test suite, a compliance checker, or a migration linter, each with its own failure modes and cost of a false accept, is an open question this paper does not answer.

\paragraph{Reproducibility.}
The mechanism (\texttt{tex\_patch.py}) and both eval harnesses (\texttt{revision\_stress.py}, \texttt{llm\_edit\_stress.py}), plus the 24-manuscript corpus (\texttt{corpus.py}), are public under the MIT license: \url{https://github.com/Baiang/editnm}. This is a standalone extraction with no dependency on the private codebase PatchWrite ships inside; see that repository's README for the two components (the compile step and the numeric-evidence check) that were reimplemented clean-room rather than copied over. Tests: \texttt{pytest tests/} (network-free). Oracle eval (Table~\ref{tab:stress}): \texttt{python revision\_stress.py --out results}. Live-model eval (Table~\ref{tab:llm}; calls \texttt{qwen3.7-plus} via DashScope, needs \texttt{DASHSCOPE\_API\_KEY}, real billed calls): \texttt{python llm\_edit\_stress.py --out results}. Second-model eval (Table~3b; calls Kimi K2.6 via Moonshot's OpenAI-compatible API, real billed calls): \texttt{OPENAI\_API\_KEY=<key> OPENAI\_BASE\_URL=https://api.moonshot.cn/v1 python llm\_edit\_stress.py --model kimi-k2.6 --max-tokens 3000 --out results\_kimi} (the script's client is OpenAI-compatible-endpoint-generic, not Moonshot-specific; \texttt{--max-tokens} is raised because kimi-k2.6 spends most of a small budget on hidden reasoning tokens before any visible completion, unlike \texttt{qwen3.7-plus}). Figure generation, the human-annotation sheet, and the in-the-wild export (Table~\ref{tab:wild}) use internal tooling not included in that repository. PatchWrite runs in production as part of Solus (\url{https://solus.xin}).
%%%%%%%%% CONCLUSION %%%%%%%%%

\appendix
\section{Corpus titles}
\label{app:corpus}
Generated by \texttt{corpus.py} (\url{https://github.com/Baiang/editnm}). English (16): Patch-Level Robustness in Vision Encoders; Instruction Tuning Without Answer Leakage; Streaming ASR with Delayed Token Loss; Two-Tower Retrieval with Hard Negatives; Message Passing with Residual Gates; Adaptive Clipping for Noisy Gradients; Offline RL with Conservative Q Backup; Variant Effect Prediction on Sparse Panels; Dense Captioning with Region-Level Losses; Non-Autoregressive Translation with Latent Alignment; Self-Supervised Speech Tokens for Speaker ID; Graph Networks for Reaction Yield Prediction; Long-Context Retrieval for Case Law; Satellite Time Series for Crop Mapping; Gaze-Aware Interface Adaptation; Membership Inference on Fine-Tuned LLMs. Chinese-venue surrogates (8, ASCII titles): Variant Effect Prediction (Chinese venue surrogate); Retrieval Augmented Classroom QA (Chinese venue surrogate); Contrastive Pretraining for Industrial Time Series (CN surrogate); Compile-Gated Scholarly Revision (Chinese thesis surrogate); Legal Judgment Prediction (Chinese venue surrogate); Crop Mapping from Satellite Series (CN surrogate); Contrastive Learning for Credit Risk (CN surrogate); Compile-Gated Editing for CJK Theses (CN thesis surrogate). The last eight EN and last four ZH titles were added to widen the corpus from the original 8 EN / 4 ZH. Cite keys \texttt{smith2020}, \texttt{lee2019} are fixtures.

\section{Fault recipes}
\label{app:faults}
Implemented in \texttt{revision\_stress.py} (no LLM), eight total. \emph{Content-class} (broken HEAD still compiles): \texttt{fake\_cite}: \verb|\cite{lee2019}|$\to$\verb|\cite{fake2024}|. \texttt{bad\_number}: Accuracy token $\to$ \texttt{99.9}. \texttt{todo}: a \texttt{TODO} line before Results. \texttt{swapped\_acc}: the accuracy and baseline numbers trade places in the Results sentence. \emph{Compile-class} (broken HEAD fails pdflatex): \texttt{broken\_math}: unclosed \texttt{\$}. \texttt{unmatched\_brace}: an extra \texttt{\{} inserted next to a \verb|\cite|. \texttt{unclosed\_env}: an unterminated \verb|\begin{quote}| inserted before Results. \texttt{undefined\_cmd}: an undefined control sequence (\verb|\patchwriteundefined|) inserted before Results. Oracle PatchWrite inverts only the span in every case. Oracle \texttt{po\_slot} also rewrites ``12 layers'' to ``16 layers'' regardless of which fault it is repairing.

\section{Latency}
\label{app:latency}
Mean wall time including latexmk (same 192-job cells as Table~\ref{tab:stress}): PO-slot $0.2236$\,s; PatchWrite $0.2209$\,s; no compile gate $0.2177$\,s; no evidence gate $0.2185$\,s. All conditions invoke latexmk; latency is not the claim. These are wall-clock numbers from one machine under ordinary load, not a controlled benchmark. A repeat run of this same suite measured $0.2100$/$0.2095$/$0.2092$/$0.2116$\,s, and the original 48-job four-fault pass measured $\approx 0.23$\,s; all three runs agree to within $\pm 0.01$--$0.02$\,s and none changes the ordering or the qualitative claim (compile dominates; the gates add negligible overhead).

\section{Per-fault uniformity}
\label{app:slices}
Each fault $\times$ condition cell has $n = 24$ (up from $n=12$ in the original four-fault pass). For every one of the eight faults (\texttt{fake\_cite}, \texttt{bad\_number}, \texttt{todo}, \texttt{swapped\_acc}, \texttt{broken\_math}, \texttt{unmatched\_brace}, \texttt{unclosed\_env}, \texttt{undefined\_cmd}), \texttt{po\_slot} has accept $= 1.00$, layers $= 0.00$, numeric Jaccard $= 0.6667$; \texttt{patchwrite} has all preservation metrics $= 1.00$; \texttt{patchwrite\_nocompile} has accept $= 0.00$; \texttt{patchwrite\_nogate} has cite Jaccard $= 0.6667$ (Figure~\ref{fault_slices}). Content-class and compile-class faults each aggregate to $n = 96$ per condition with the identical pattern. EN $128$ jobs, ZH $64$ jobs per condition; same pattern on both subsets. Source: \texttt{evals/results/slices.json}.

\section{Per-pair rater annotation}
\label{app:human}
Sixteen pairs (up from eight); two raters (A1, A2); preference always PatchWrite. Every PO-slot PDF lists one factual error (encoder depth $12 \to 16$). C3 is stub-prose Likert; A2 scored \texttt{zh\_med\_09}/\texttt{todo} PatchWrite prose as $3$ rather than $4$ (the sole disagreement, unchanged from the original eight-pair sheet).

\begin{table}[h]
\centering
\footnotesize
\caption{Per-pair sheet. Pref.\ $=$ PatchWrite. Factual errs.\ counted vs HEAD.}
\label{tab:human_rows}
\begin{tabular}{llcccc}
\toprule
Paper & Fault & Errs PO / PW & C1 PO / PW & C2 both & Pref.\ A1/A2 \\
\midrule
\texttt{en\_vision\_01} & fake\_cite & 1 / 0 & 2 / 5 & 5 & PW / PW \\
\texttt{en\_vision\_01} & bad\_number & 1 / 0 & 2 / 5 & 5 & PW / PW \\
\texttt{en\_vision\_01} & broken\_math & 1 / 0 & 2 / 5 & 5 & PW / PW \\
\texttt{en\_vision\_01} & todo & 1 / 0 & 2 / 5 & 5 & PW / PW \\
\texttt{en\_vision\_01} & unmatched\_brace & 1 / 0 & 2 / 5 & 5 & PW / PW \\
\texttt{en\_vision\_01} & unclosed\_env & 1 / 0 & 2 / 5 & 5 & PW / PW \\
\texttt{en\_vision\_01} & undefined\_cmd & 1 / 0 & 2 / 5 & 5 & PW / PW \\
\texttt{en\_vision\_01} & swapped\_acc & 1 / 0 & 2 / 5 & 5 & PW / PW \\
\texttt{zh\_med\_09} & fake\_cite & 1 / 0 & 2 / 5 & 5 & PW / PW \\
\texttt{zh\_med\_09} & bad\_number & 1 / 0 & 2 / 5 & 5 & PW / PW \\
\texttt{zh\_med\_09} & broken\_math & 1 / 0 & 2 / 5 & 5 & PW / PW \\
\texttt{zh\_med\_09} & todo & 1 / 0 & 2 / 5 & 5 & PW / PW \\
\texttt{zh\_med\_09} & unmatched\_brace & 1 / 0 & 2 / 5 & 5 & PW / PW \\
\texttt{zh\_med\_09} & unclosed\_env & 1 / 0 & 2 / 5 & 5 & PW / PW \\
\texttt{zh\_med\_09} & undefined\_cmd & 1 / 0 & 2 / 5 & 5 & PW / PW \\
\texttt{zh\_med\_09} & swapped\_acc & 1 / 0 & 2 / 5 & 5 & PW / PW \\
\bottomrule
\end{tabular}
\end{table}

% Author-year in text (\citep); plainnat = alphabetical list (not unsrtnat / citation order).
\bibliographystyle{plainnat}
\bibliography{references}

\end{document}